\def\year{2019}\relax
\documentclass[letterpaper]{article} 
\usepackage{aaai19}  
\usepackage{times}  
\usepackage{helvet}  
\usepackage{courier}  
\usepackage{url}  
\usepackage{graphicx}  
\usepackage{booktabs}
\usepackage{amsmath}
\usepackage{amssymb}
\usepackage{bm}
\usepackage{algorithm}
\usepackage{algorithmic}
\usepackage{color}
\usepackage{multirow}
\usepackage{subfig}
\newcommand{\argmax}{\mathop{\rm argmax}\limits}

\newcommand{\Tref}[1]{Table~\ref{#1}}
\newcommand{\Eref}[1]{Eq.~(\ref{#1})}
\newcommand{\Fref}[1]{Fig.~\ref{#1}}

\graphicspath{{./figs/}}

\begin{document}
%
\title{Fully Convolutional Network with Multi-Step Reinforcement Learning\\for Image Processing}
\author{Ryosuke Furuta, Naoto Inoue, Toshihiko Yamasaki\\
Department of Information and Communication Engineering, The University of Tokyo, Tokyo, Japan\\
\{furuta, inoue, yamasaki\}@hal.t.u-tokyo.ac.jp
}
\maketitle
\begin{abstract}
This paper tackles a new problem setting: reinforcement learning with pixel-wise rewards ({\it pixelRL}) for image processing.
After the introduction of the deep Q-network, deep RL has been achieving great success.
However, the applications of deep RL for image processing are still limited.
Therefore, we extend deep RL to pixelRL for various image processing applications.
In pixelRL, each pixel has an agent, and the agent changes the pixel value by taking an action.
We also propose an effective learning method for pixelRL that significantly improves the performance by considering not only the future states of the own pixel but also those of the neighbor pixels.
The proposed method can be applied to some image processing tasks that require pixel-wise manipulations, where deep RL has never been applied.

We apply the proposed method to three image processing tasks: image denoising, image restoration, and local color enhancement.
Our experimental results demonstrate that the proposed method achieves comparable or better performance, compared with the state-of-the-art methods based on supervised learning.
\end{abstract}

\section{Introduction}\label{sec:intro}
After the introduction of the deep Q-network (DQN)~\cite{mnih2013playing}, which can play Atari games on the human level, much attention has been focused on deep reinforcement learning (RL).
Recently, deep RL is also applied to a variety of image processing tasks~\cite{li2017a2,shuyue2018ffnet,jongchan2018distort}.
However, these methods can execute only global actions for the entire image and are limited to simple applications, e.g., image cropping~\cite{li2017a2} and global color enhancement~\cite{jongchan2018distort,hu2017exposure}.
Therefore, these methods cannot be applied to applications that require pixel-wise manipulations such as image denoising.

To overcome this drawback, we propose a new problem setting: pixelRL for image processing.
PixelRL is a multi-agent RL problem, where the number of agents is equal to that of pixels.
The agents learn the optimal behavior to maximize the mean of the expected total rewards at all pixels.
Each pixel value is regarded as the current state and is iteratively updated by the agent's action.
Applying the existing techniques of the multi-agent RL to pixelRL is impractical in terms of computational cost because the number of agents is extremely large.
Therefore, we solve the problem by employing the fully convolutional network (FCN).
The merit of using FCN is that all the agents can share the parameters and learn efficiently.
Herein, we also propose {\it reward map convolution}, which is an effective learning method for pixelRL.
By the proposed reward map convolution, each agent considers not only the future states of its own pixel but also those of the neighbor pixels.

The proposed pixelRL is applied to image denoising, image restoration, and local color enhancement.
To the best of our knowledge, this is the first work to apply RL to such low-level image processing for each pixel or each local region.
Our experimental results show that the agents trained with the pixelRL and the proposed reward map convolution achieve comparable or better performance, compared with state-of-the-art methods based on supervised learning.
Although the actions must be pre-defined for each application, the proposed method is interpretable by observing the actions executed by the agents, which is a novel and different point from the existing deep learning-based image processing methods for such applications.

Our contributions are summarized as follows:
\begin{itemize}
\item We propose a novel problem setting: pixelRL for image processing, where the existing techniques for multi-agents RL cannot be applied. 
\item We propose {\it reward map convolution}, which is an effective learning method for pixelRL and boosts the performance.
\item We apply the pixelRL to image denoising, image restoration, and local color enhancement.
The proposed method is a completely novel approach for these tasks, and shows better or comparable performance, compared with state-of-the-art methods.
\item The actions executed by the agents are interpretable to humans, which is of great difference from conventional CNNs.
\end{itemize}

\section{Related Works}\label{sec:rel}
\subsection{Deep RL for image processing}
Very recently, deep RL has been used for some image processing applications.
Cao et al.~\shortcite{cao2017attention} proposed a super-resolution method for face images.
The agent first chooses a local region and inputs it to the local enhancement network.
The enhancement network converts the local patch to a high-resolution one, and the agents chooses the next local patch that should be enhanced.
This process is repeated until the maximum time step; consequently, the entire image is enhanced.
Li et al.~\shortcite{li2017a2} used deep RL for image cropping.
The agent iteratively reshapes the cropping window to maximize the aesthetics score of the cropped image.
Yu et al.~\shortcite{yu2018crafting} proposed the RL-restore method, where the agent selects a toolchain from a toolbox (a set of light-weight CNNs) to restore a corrupted image.
Park et al.~\shortcite{jongchan2018distort} proposed a color enhancement method using DQN.
The agent iteratively chooses the image manipulation action (e.g., increase brightness) and retouches the input image.
The reward is defined as the negative distance between the retouched image by the agent and the one by an expert.
A similar idea is proposed by Hu et al.~\shortcite{hu2017exposure}, where the agent retouches from RAW images.
As discussed in the introduction, all the above methods execute global actions for entire images.
In contrast, we tackle pixelRL, where pixel-wise actions can be executed.

Wulfmeier et al.~\shortcite{wulfmeier2015maximum} used the FCN to solve the inverse reinforcement learning problem.
This problem setting is different from ours because one pixel corresponds to one state, and the number of agents is one in their setting.
In contrast, our pixelRL has one agent at each pixel.

\subsection{Image Denoising}
Image denoising methods are classified into two categories: non-learning and learning based.
Many classical methods are categorized into the former class (e.g., BM3D~\cite{dabov2007image}).
Although learning-based methods include dictionary-based methods such as~\cite{mairal2009non}, the recent trends in image denoising is neural network-based methods~\cite{zhang2017learning,lefkimmiatis2017non}.
Generally, neural-network-based methods have shown better performances, compared with non-leaning-based methods.

Our denoising method based on pixelRL is a completely different approach from other neural network-based methods.
While most of neural-network-based methods learn to regress noise or true pixel values from a noisy input, our method iteratively removes noise with the sequence of simple pixel-wise actions (basic filters).

\subsection{Image Restoration}
Similar to image denoising, image restoration (also called image inpainting) methods are divided into non-learning and learning-based methods.
In the former methods such as~\cite{bertalmio2000image}, the target blank regions are filled by propagating the pixel values or gradient information around the regions.
The filling process is highly sophisticated, but they are based on a handcrafted-algorithm.
Roth and Black~\shortcite{roth2005fields} proposed a Markov random field-based model to learn the image prior to the neighbor pixels.
Mairal et al.~\shortcite{mairal2008sparse} proposed a learning-based method that creates a dictionary from an image database using K-SVD, and applied it to image  denoising and inpainting.
Recently, deep-neural-network-based methods were proposed~\cite{xie2012image,DBLP:journals/corr/abs-1712-09078}, and the U-Net-based inpainting method~\cite{DBLP:journals/corr/abs-1712-09078} showed much better performance than other methods.

Our method is categorized into the learning-based method because we used training images to optimize the policies.
However, similar to the classical inpainting methods, our method successfully propagates the neighbor pixel values with the sequence of basic filters.

\subsection{Color Enhancement}
One of the classical methods is color transfer proposed by Reinhard et al.~\shortcite{reinhard2001color}, where the global color distribution of the reference image is transfered to the target image.
Hwang et al.~\shortcite{hwang2012context} proposed an automatic local color enhancement method based on image retrieval.
This method enhances the color of each pixel based on the retrieved images with smoothness regularization, which is formulated as a Gaussian MRF optimization problem.

Yan et al.~\shortcite{yan2016automatic} proposed the first color enhancement method based on deep learning.
They used a DNN to learn a mapping function from the carefully designed pixel-wise features to the desired pixel values.
Gharbi et al.~\shortcite{gharbi2017deep} used a CNN as a trainable bilateral filter for high-resolution images and applied it to some image processing tasks.
Similarly, for fast image processing, Chen et al.~\shortcite{chen2017fast} adopted an FCN to learn an approximate mapping from the input to the desired images.
Unlike deep learning-based methods that learn the input for an output mapping, our color enhancement method is interpretable because our method enhances each pixel value iteratively with actions such as~\cite{jongchan2018distort,hu2017exposure}.

\section{Background Knowledge}
Herein, we extend the asynchronous advantage actor-critic (A3C)~\cite{mnih2016asynchronous} for the pixelRL problem because A3C showed good performance with efficient training in the original paper\footnote{Note that we can employ any deep RL methods such as DQN instead of A3C.}.
In this section, we briefly review the training algorithm of A3C.
A3C is one of the actor-critic methods, which has two networks: policy network and value network.
We denote the parameters of each network as $\theta_p$ and $\theta_v$, respectively.
Both networks use the current state $s^{(t)}$ as the input, where $s^{(t)}$ is the state at time step $t$.
The value network outputs the value $V(s^{(t)})$: the expected total rewards from state $s^{(t)}$, which shows how good the current state is.
The gradient for $\theta_v$ is computed as follows:
\begin{multline}
R^{(t)}=r^{(t)}+\gamma r^{(t+1)}+\gamma^2 r^{(t+2)}+\cdots \\
+\gamma^{n-1}r^{(t+n-1)}+\gamma^n V(s^{(t+n)}),\label{eq:R}
\end{multline}
\begin{equation}
d\theta_v=\nabla_{\theta_v}\left(R^{(t)}-V(s^{(t)})\right)^2,
\end{equation}
where $\gamma^i$ is the $i$-th power of the discount factor $\gamma$.

The policy network outputs the policy $\pi(a^{(t)}|s^{(t)})$ (probability through softmax) of taking action $a^{(t)}\in \mathcal{A}$.
Therefore, the output dimension of the policy network is $|\mathcal{A}|$.
The gradient for $\theta_p$ is computed as follows:
\begin{eqnarray}
A(a^{(t)},s^{(t)})&=&R^{(t)}-V(s^{(t)}),\label{eq:original_adv} \\ 
d\theta_p&=&-\nabla_{\theta_p}\log \pi(a^{(t)}|s^{(t)})A(a^{(t)},s^{(t)}).
\end{eqnarray}
$A(a^{(t)},s^{(t)})$ is called the advantage, and $V(s^{(t)})$ is subtracted in~\Eref{eq:original_adv} to reduce the variance of the gradient.
For more details, see~\cite{mnih2016asynchronous}.

\section{Reinforcement Learning with Pixel-wise Rewards (PixelRL)}
Here, we describe the proposed pixelRL problem setting.
Let $I_i$ be the $i$-th pixel in the input image $\bm{I}$ that has $N$ pixels $(i=1,\cdots,N)$.
Each pixel has an agent, and its policy is denoted as $\pi_i(a_i^{(t)}|s_i^{(t)})$, where $a_i^{(t)} (\in \mathcal{A})$ and $s_i^{(t)}$ are the action and the state of the $i$-th agent at time step $t$, respectively.
$\mathcal{A}$ is the pre-defined action set, and $s_i^{(0)}=I_i$. 
The agents obtain the next states $\bm{s}^{(t+1)}=(s_1^{(t+1)},\cdots,s_N^{(t+1)})$ and rewards $\bm{r}^{(t)}=(r_1^{(t)},\cdots,r_N^{(t)})$ from the environment by taking the actions $\bm{a}^{(t)}=(a_1^{(t)},\cdots,a_N^{(t)})$.
The objective of the pixelRL problem is to learn the optimal policies $\bm{\pi}=(\pi_1,\cdots,\pi_N)$ that maximize the mean of the total expected rewards at all pixels:
\begin{eqnarray}
\bm{\pi}^*&=&\argmax_{\bm{\pi}}E_{\bm{\pi}}\left(\sum_{t=0}^{\infty}\gamma^t \overline{r}^{(t)}\right),\\
\overline{r}^{(t)}&=&\frac{1}{N}\sum_{i=1}^N r^{(t)}_i,
\end{eqnarray}
where $\overline{r}^{(t)}$ is the mean of the rewards $r_i^{(t)}$ at all pixels.

A naive solution for this problem is to train a network that output Q-values or policies for all possible set of actions $\bm{a}^{(t)}$.
However, it is computationally impractical because the dimension of the last fully connected layer must be $|\mathcal{A}|^N$, which is too large.

Another solution is to divide this problem into $N$ independent subproblems and train $N$ networks, where we train the $i$-th agent to maximize the expected total reward at the $i$-th pixel:
\begin{equation}
\pi_i^*=\argmax_{\pi_i}E_{\pi_i}\left(\sum_{t=0}^{\infty}\gamma^t r_i^{(t)} \right).
\end{equation}
However, training $N$ networks is also computationally impractical when the number of pixels is large.
In addition, it treats only the fixed size of images.
To solve the problems, we employ a FCN instead of $N$ networks.
By using the FCN, all the $N$ agents can share the parameters, and we can parallelize the computation of $N$ agents on a GPU, which renders the training efficient.
Herein, we employ A3C and extend it to the fully convolutional form.
Our architecture is illustrated in the supplemental material.

The pixelRL setting is different from typical multi-agent RL problems in terms of two points.
The first point is that the number of agents $N$ is extremely large ($>10^5$).
Therefore, typical multi-agent learning techniques such as~\cite{lowe2017multi} cannot be directly applied to the pixelRL.
Next, the agents are arrayed in a 2D image plane.
In the next section, we propose an effective learning method that boosts the performance of the pixelRL agents by leveraging this property, named {\it reward map convolution}.

\section{Reward Map Convolution}\label{sec:rconv}
Here, for the ease of understanding, we first consider the one-step learning case (i.e., $n=1$ in~\Eref{eq:R}).

When the receptive fields of the FCNs are 1x1 (i.e., all the convolution filters in the policy and value network are 1x1), the $N$ subproblems are completely independent.
In that case, similar to the original A3C, the gradient of the two networks are computed as follows:
\begin{eqnarray}
R_i^{(t)}&=&r_i^{(t)}+\gamma V(s_i^{(t+1)}),\label{eq:ith_R}\\
d\theta_v&=&\nabla_{\theta_v}\frac{1}{N}\sum_{i=1}^N\left(R_i^{(t)}-V(s_i^{(t)})\right)^2, \label{eq:ith_gradv}\\
A(a_i^{(t)},s_i^{(t)})&=&R_i^{(t)}-V(s_i^{(t)}), \label{eq:i_adv}
\end{eqnarray}
\begin{equation}
d\theta_p=-\nabla_{\theta_p}\frac{1}{N}\sum_{i=1}^N\log \pi(a_i^{(t)}|s_i^{(t)})A(a_i^{(t)},s_i^{(t)}).\label{eq:ith_gradp}
\end{equation}
As shown in~Eqs.~(\ref{eq:ith_gradv}) and (\ref{eq:ith_gradp}), the gradient for each network parameter is the average of the gradients at all pixels.

However, one of the recent trends in CNNs is to enlarge the receptive field to boost the network performance~\cite{yu2017dilated,zhang2017learning}.
Our network architecture, which was inspired by~\cite{zhang2017learning} in the supplemental material, has a large receptive field.
In this case, the policy and value networks observe not only the $i$-th pixel $s_i^{(t)}$ but also the neighbor pixels to output the policy $\pi$ and value $V$ at the $i$-th pixel.
In other words, the action $a_i^{(t)}$ affects not only the $s_i^{(t+1)}$ but also the policies and values in $\mathcal{N}(i)$ at the next time step, where $\mathcal{N}(i)$ is the local window centered at the $i$-th pixel.
Therefore, to consider it, we replace $R_i$ in~\Eref{eq:ith_R} as follows:
\begin{equation}
R_i^{(t)}=r_i^{(t)}+\gamma \sum_{j\in \mathcal{N}(i)}w_{i-j} V(s_j^{(t+1)}),\label{eq:ith_R_conv}
\end{equation}
where $w_{i-j}$ is the weight that means how much we consider the values $V$ of the neighbor pixels at the next time step ($t+1$).
$\bm{w}$ can be regarded as a convolution filter weight and can be learned simultaneously with the network parameters $\theta_p$ and $\theta_v$.
It is noteworthy that the second term in~\Eref{eq:ith_R_conv} is a 2D convolution because each pixel $i$ has a 2D coordinate $(i_x,i_y)$.

Using the matrix form, we can define the $\bm{R}^{(t)}$ in the $n$-step case.
\begin{multline}
\bm{R}^{(t)}=\bm{r}^{(t)}+\gamma \bm{w}*\bm{r}^{(t+1)}+\gamma^2 \bm{w}^2*\bm{r}^{(t+2)}+\cdots \\
+\gamma^{n-1}\bm{w}^{n-1}*\bm{r}^{(t+n-1)}+\gamma^n\bm{w}^n*V(\bm{s}^{(t+n)}),\label{eq:nstep_R_conv}
\end{multline}
where $*$ is the convolution operator, and $\bm{w}^{n}*\bm{r}$ denotes the $n$-times convolution on $\bm{r}$ with the filter $\bm{w}$.
Similar to $\theta_p$ and $\theta_v$ in~Eqs.~(\ref{eq:ith_gradv}) and (\ref{eq:ith_gradp}), the gradient for $\bm{w}$ is computed as follows:
\begin{multline}
d\bm{w}=-\nabla_{\bm{w}}\frac{1}{N}\sum_{i=1}^N\log\pi(a_i^{(t)}|s_i^{(t)})(R_i^{(t)}-V(s_i^{(t)}))\\
+\nabla_{\bm{w}}\frac{1}{N}\sum_{i=1}^N(R_i^{(t)}-V(s_i^{(t)}))^2.\label{eq:gradw}
\end{multline}
Similar to typical policy gradient algorithms, the first term in~\Eref{eq:gradw} encourages a higher expected total reward.
The second term operates as a regularizer such that $R_i$ is not deviated from the prediction $V(s_i^{(t)})$ by the convolution.

\section{Applications and Results}
We implemented the proposed method on Python with Chainer \cite{tokui2015chainer} and ChainerRL~\footnote{https://github.com/chainer/chainerrl} libraries, and applied it to three different applications.

\subsection{Image denoising}
\subsubsection{Method}
\begin{table}[t]
\caption{Actions for image denoising and restoration.}
\centering
{
{\small
  \begin{tabular}{cccc} \toprule
      & action & filter size & parameter \\ \toprule
     1 & box filter & 5x5 & - \\
     2 & bilateral filter & 5x5 & $\sigma_c=1.0, \sigma_S=5.0$ \\
     3 & bilateral filter & 5x5 & $\sigma_c=0.1, \sigma_S=5.0$ \\
     4 & median filter & 5x5 & - \\
     5 & Gaussian filter & 5x5 & $\sigma=1.5$ \\
     6 & Gaussian filter & 5x5 & $\sigma=0.5$ \\
     7 & pixel value += 1 & - & - \\
     8 & pixel value -= 1 & - & - \\
     9 & do nothing & - & - \\
 \bottomrule
  \end{tabular} 
}
}
\label{tbl:actions_denoise}
\end{table}
The input image $\bm{I}(=\bm{s}^{(0)})$ is a noisy gray scale image, and the agents iteratively remove the noises by executing actions. 
It is noteworthy that the proposed method can also be applied to color images by independently manipulating on the three channels.
\Tref{tbl:actions_denoise} shows the list of actions that the agents can execute, which were empirically decided.
We defined the reward $r_i^{(t)}$ as follows:
\begin{equation}
r_i^{(t)}=(I_i^{target}-s_i^{(t)})^2-(I_i^{target}-s_i^{(t+1)})^2,\label{eq:reward_denoise}
\end{equation}
where $I_i^{target}$ is the $i$-th pixel value of the original clean image.
Intuitively, \Eref{eq:reward_denoise} means how much the squared error on the $i$-th pixel was decreased by the action $a_i^{(t)}$.
As shown in~\cite{maes2009structured}, maximizing the total reward in~\Eref{eq:reward_denoise} is equivalent to minimizing the squared error between the final state $\bm{s}^{(t_{max})}$ and the original clean image $\bm{I}^{target}$.
We set the number of training episodes to 30,000 and the length of each episode $t_{max}$ to 5.
We set the filter size of $\bm{w}$ to $33\times 33$, which is equal to the receptive field size of the policy and value networks.
The other implementation details are shown in the supplemental material. 
We used BSD68 dataset~\cite{roth2005fields}, which has 428 train images and 68 test images.
Similar to~\cite{zhang2017learning}, we added 4,774 images from Waterloo exploration database~\cite{ma2017waterloo} to the training set.

\subsubsection{Results}\label{sec:denoise_results}
\begin{table}[t]
\caption{PSNR [dB] on BSD68 test set with Gaussian noise.}
\centering
{
{\small
\scalebox{0.9}{
  \begin{tabular}{ccc|ccc} \toprule
     \multicolumn{3}{c|}{\multirow{2}{*}{Method}} & \multicolumn{3}{c}{std. $\sigma$}\\
      & & & 15 & 25 & 50 \\ \toprule
     \multicolumn{3}{c|}{BM3D~\cite{dabov2007image}} & 31.07 & 28.57 & 25.62 \\
     \multicolumn{3}{c|}{WNNM~\cite{gu2014weighted}} & 31.37 & 28.83 & 25.87 \\
     \multicolumn{3}{c|}{TNRD~\cite{chen2017trainable}} & 31.42 & 28.92 & 25.97 \\
     \multicolumn{3}{c|}{MLP~\cite{burger2012image}} & - & 28.96 & 26.03 \\
     \multicolumn{3}{c|}{CNN~\cite{zhang2017learning}} & 31.63 & 29.15 & 26.19 \\
     \multicolumn{3}{c|}{CNN~\cite{zhang2017learning} +aug.} & {\bf 31.66} & {\bf 29.18} & {\bf 26.20} \\ \midrule
     \multicolumn{3}{c|}{Proposed} & & & \\
     +convGRU & +RMC & +aug. & & & \\
      & & & 31.17 & 28.75 & 25.78 \\
      \checkmark& & & 31.26 & 28.83 & 25.87 \\
      \checkmark&\checkmark & & 31.40 & 28.85 & 25.88 \\
      \checkmark&\checkmark &\checkmark & 31.49 & 28.94 & 25.95 \\
 \bottomrule
  \end{tabular} 
}
}
}
\label{tbl:comp_gaussian}
\end{table}
\Tref{tbl:comp_gaussian} shows the comparison of Gaussian denoising with other methods.
RMC is the abbreviation for reward map convolution.
Aug. means the data augmentation for test images, where a single test image was augmented to eight images by a left-right flip and $90^{\circ}$, $180^{\circ}$, and $270^{\circ}$ rotations, similar to~\cite{timofte2017ntire}.
We observed that CNN~\cite{zhang2017learning} is the best.
However, the proposed method achieved the comparable results with other state-of-the-art methods.
Adding the convGRU to the policy network improved the PSNR by approximately +0.1dB.
The RMC significantly improved the PSNR when $\sigma=15$, but improved little when $\sigma=25$ and $50$.
That is because the agents can obtain much reward by removing the noises at their own pixels rather than considering the neighbor pixels when the noises are strong.
The augmentation for test images further boosted the performance.
We report the CNN~\cite{zhang2017learning} with the same augmentation for a fair comparison.
Almost similar results were obtained on Poisson denoising, which are shown in the supplemental material.

\begin{figure}[t]
    \begin{center}
        \includegraphics[width=0.9\linewidth]{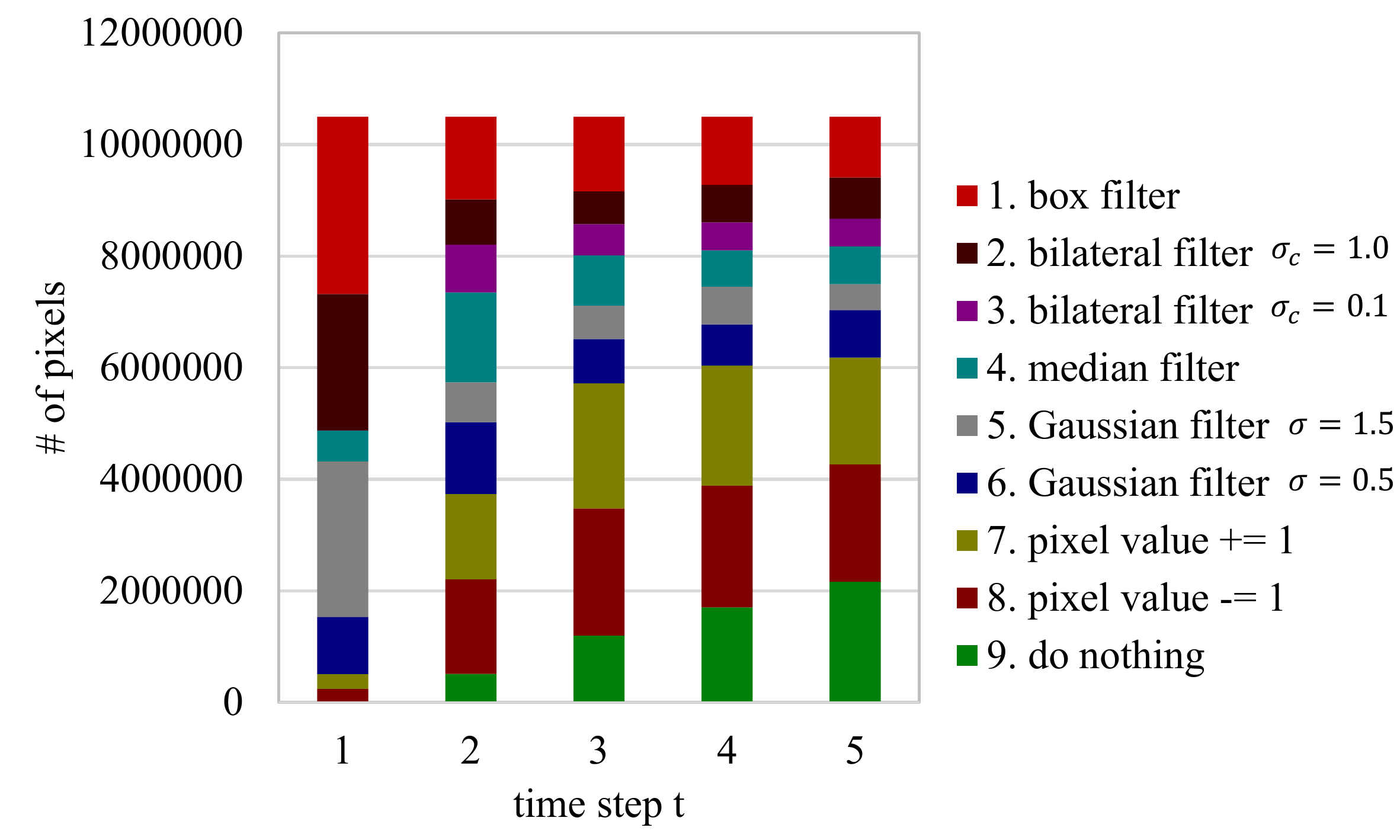}
        \caption{Number of actions executed at each time step for Gaussian denoising ($\sigma=50$) on the BSD68 test set.}
        \label{fig:actions_denoise}
    \end{center}    
\end{figure}
\Fref{fig:actions_denoise} shows the number of actions executed by the proposed method at each time step for Gaussian denoising ($\sigma=50$) on the BSD68 test set.
We observed that the agents successfully obtained a strategy in which they first removed the noises using strong filters (box filter, bilateral filter $\sigma_c=1.0$, and Gaussian filter $\sigma=1.5$); subsequently they adjusted the pixel values by the other actions (pixel values +=1 and -=1).

\begin{table}[t]
\caption{PSNR [dB] on BSD68 test set with Salt\&Pepper noise.}
\centering
{
{\small
  \begin{tabular}{ccc|ccc} \toprule
     \multicolumn{3}{c|}{\multirow{2}{*}{Method}} & \multicolumn{3}{c}{Noise density}\\
      & & & 0.1 & 0.5 & 0.9 \\ \toprule
     \multicolumn{3}{c|}{CNN~\cite{zhang2017learning}} & 40.16 & 29.19 & 23.58 \\
     \multicolumn{3}{c|}{CNN~\cite{zhang2017learning} +aug.} & {\bf 40.40} & 29.40 & 23.76 \\ \midrule
     \multicolumn{3}{c|}{Proposed} & & & \\
     +convGRU & +RMC & +aug. & & & \\
      & & & 36.51 & 27.91 & 22.73 \\
      \checkmark& & & 37.86 & 29.26 & 23.54 \\
      \checkmark&\checkmark & & 38.46 & 29.78 & 23.78 \\
      \checkmark&\checkmark &\checkmark & 38.82 & {\bf 29.92} & {\bf 23.81} \\
 \bottomrule
  \end{tabular} 
}
}
\label{tbl:comp_sp}
\end{table}
\Tref{tbl:comp_sp} shows the comparison of salt and pepper denoising.
We observed that the RMC significantly improved the performance.
In addition, the proposed method outperformed the CNN~\cite{zhang2017learning} when the noise density is 0.5 and 0.9.
Unlike Gaussian and Poisson noises, it is difficult to regress the noise with CNN when the noise density is high because the information of the original pixel value is lost (i.e., the pixel value was changed to 0 or 255 by the noise).
In contrast, the proposed method can predict the true pixel values from the neighbor pixels with the iterative filtering actions.

\begin{figure}[t]
    \begin{center}
        \includegraphics[width=\linewidth]{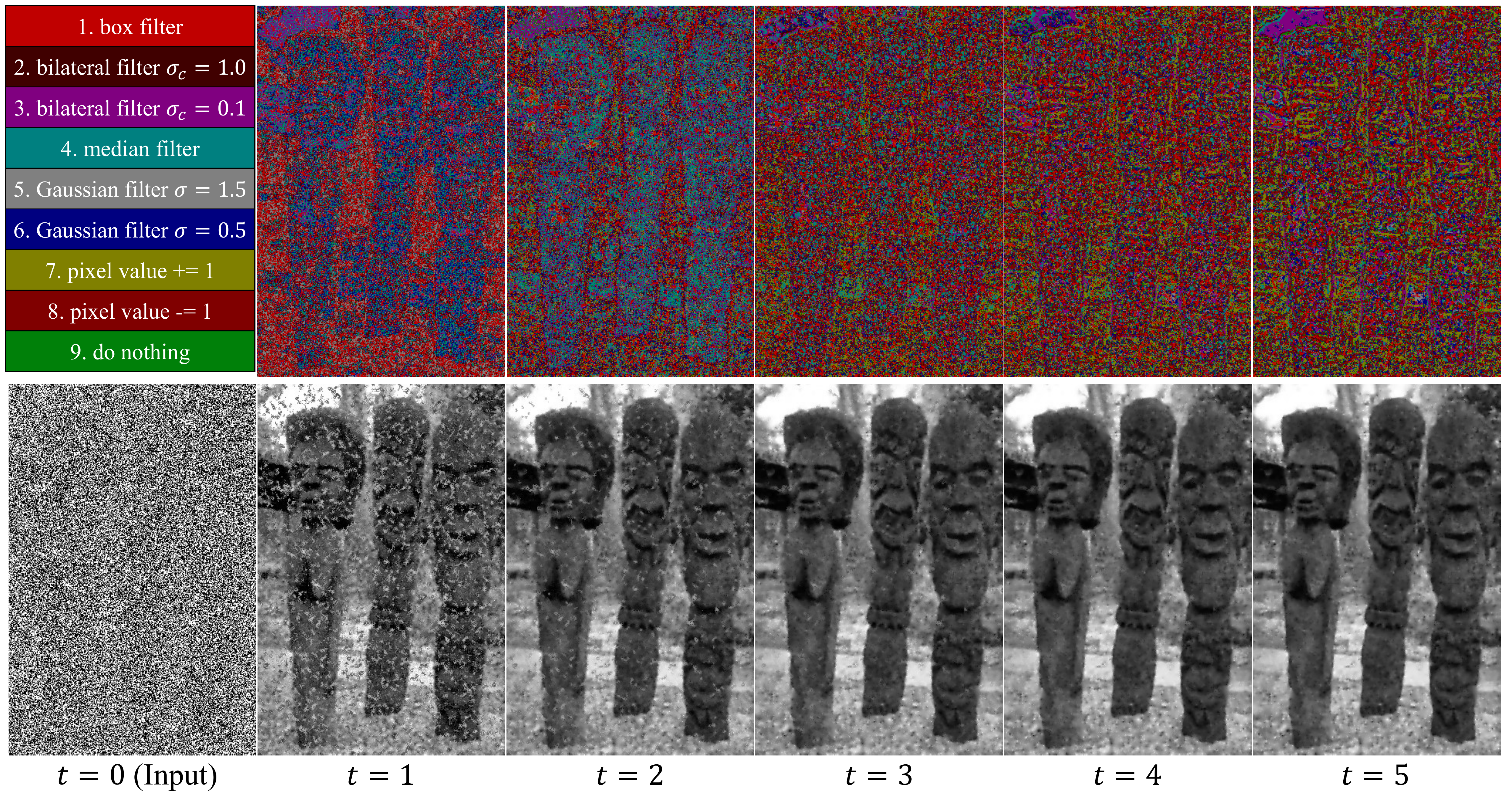}
        \caption{Denoising process of the proposed method and the action map at each time step for salt and pepper denoising (density=0.9).}
        \label{fig:vis_denoise}
    \end{center}    
\end{figure}
We visualize the denoising process of the proposed method, and the action map at each time step in~\Fref{fig:vis_denoise}.
We observed that the noises are iteratively removed by the chosen actions.

\begin{figure}[t]
    \begin{center}
        \includegraphics[width=1.0\linewidth]{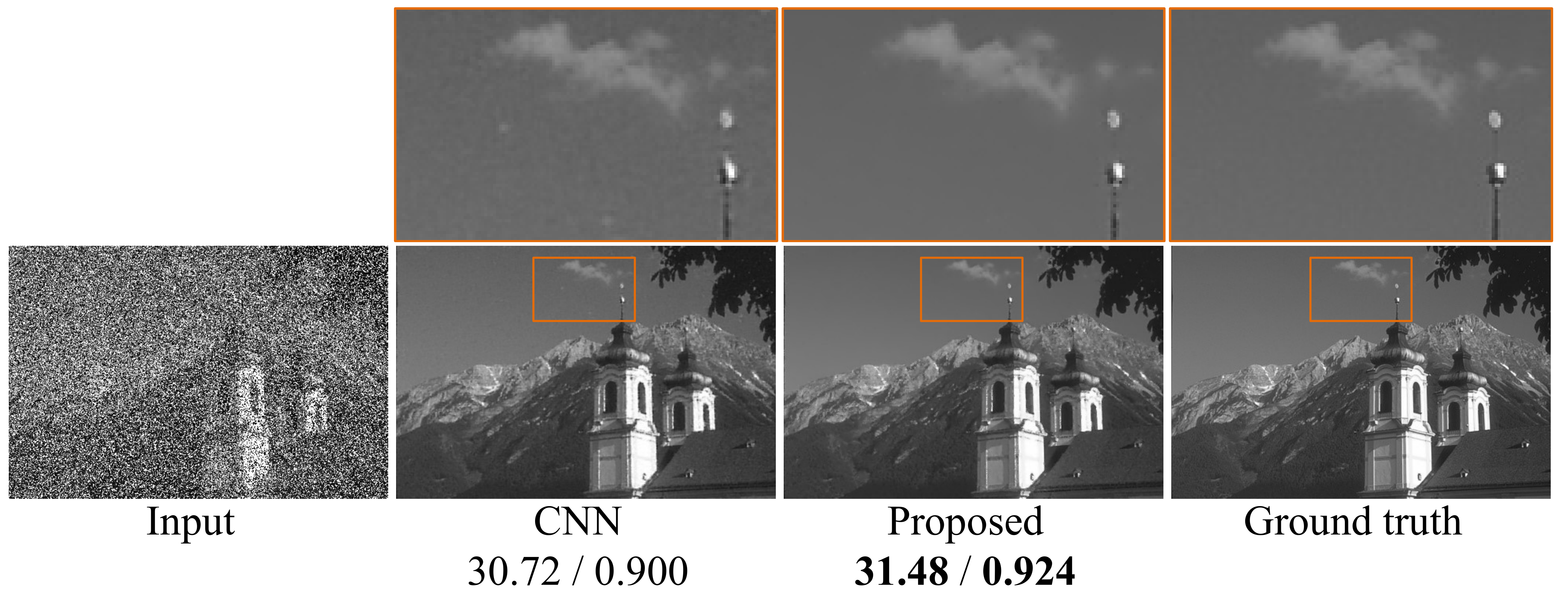}
        \caption{Qualitative comparison of the proposed method and CNN~\cite{zhang2017learning} for salt and pepper noise (density=0.5). PSNR / SSIM are reported.}
        \label{fig:qualitative_denoise}
    \end{center}    
\end{figure}
\Fref{fig:qualitative_denoise} shows the qualitative comparison with CNN~\cite{zhang2017learning}.
The proposed method achieved both quantitative and visually better results for salt and pepper denoising.

\subsection{Image Restoration}\label{sec:rest}
\subsubsection{Method}
We applied the proposed method to ``blind'' image restoration, where no mask of blank regions is provided.
The proposed method iteratively inpaints the blank regions by executing actions.
We used the same actions and reward function as those of image denoising, which are shown in~\Tref{tbl:actions_denoise} and~\Eref{eq:reward_denoise}, respectively.

For training, we used 428 training images from the BSD68 train set, 4,774 images from the Waterloo exploration database, and 20,122 images from the ILSVRC2015 val set~\cite{ILSVRC15} (a total of 25,295 images).
We also used 11,343 documents from the Newsgroups 20 train set~\cite{lang1995newsweeder}.
During the training, we created each training image by randomly choosing an image from the 25,295 images and a document from 11,343 documents, and overlaid it on the image.
The font size was randomly decided from the range [10,30].
The font type was randomly chosen between {\it Arial} and {\it Times New Roman}, where the bold and Italic options were randomly added.
The intensity of the text region was randomly chosen from 0 or 255.
We created the test set that has 68 images by overlaying the randomly chosen 68 documents from the Newsgroup 20 test set on the BSD68 test images.
The settings of the font size and type were the same as those of the training.
The random seed for the test set was fixed between the different methods.
All the hyperparameters were same as those in image denoising, except for the length of the episodes, i.e., $t_{max}=15$.

\subsubsection{Results}\label{sec:restoration_results}
\begin{table}[t]
\caption{Comparison on image restoration.}
\centering
{
{\small
  \begin{tabular}{cc|cc} \toprule
     \multicolumn{2}{c|}{Method} & PSNR [dB] & SSIM \\ \toprule
     \multicolumn{2}{c|}{Net-D and Net-E~\cite{DBLP:journals/corr/abs-1712-09078}} & 29.53 & 0.846 \\
     \multicolumn{2}{c|}{CNN~\cite{zhang2017learning}} & 29.75 & 0.858 \\ \midrule
     \multicolumn{2}{c|}{Proposed} & \\
     +convGRU & +RMC & & \\
      \checkmark& & 29.50 & 0.858 \\
      \checkmark&\checkmark & {\bf 29.97} & {\bf 0.868} \\
 \bottomrule
  \end{tabular} 
}
}
\label{tbl:comp_rest}
\end{table}
\Tref{tbl:comp_rest} shows the comparison of the averaged PSNR between the output and ground-truth images.
We saved the models of the compared methods at every one epoch, and reported the best results.
For the proposed method, we saved the model at every 300 episodes ($\simeq$ 0.76 epoch) and reported the best results.
Here, we compared the proposed method with the two methods (Net-E and Net-D~\cite{DBLP:journals/corr/abs-1712-09078} and CNN~\cite{zhang2017learning}) because the Net-E and Net-D achieved much better results than the other restoration methods in the original paper.
We found that the RMC significantly improved the performance, and the proposed method obtained the best result.
This is the similar reason to the case of the salt and pepper noise.
Because the information of the original pixel value is lost by the overlaid texts, its regression is difficult.
In contrast, the proposed method predicts the true pixel value by iteratively propagating the neighbor pixel values with the filtering actions.

\begin{figure}[t]
    \begin{center}
        \includegraphics[width=1.0\linewidth]{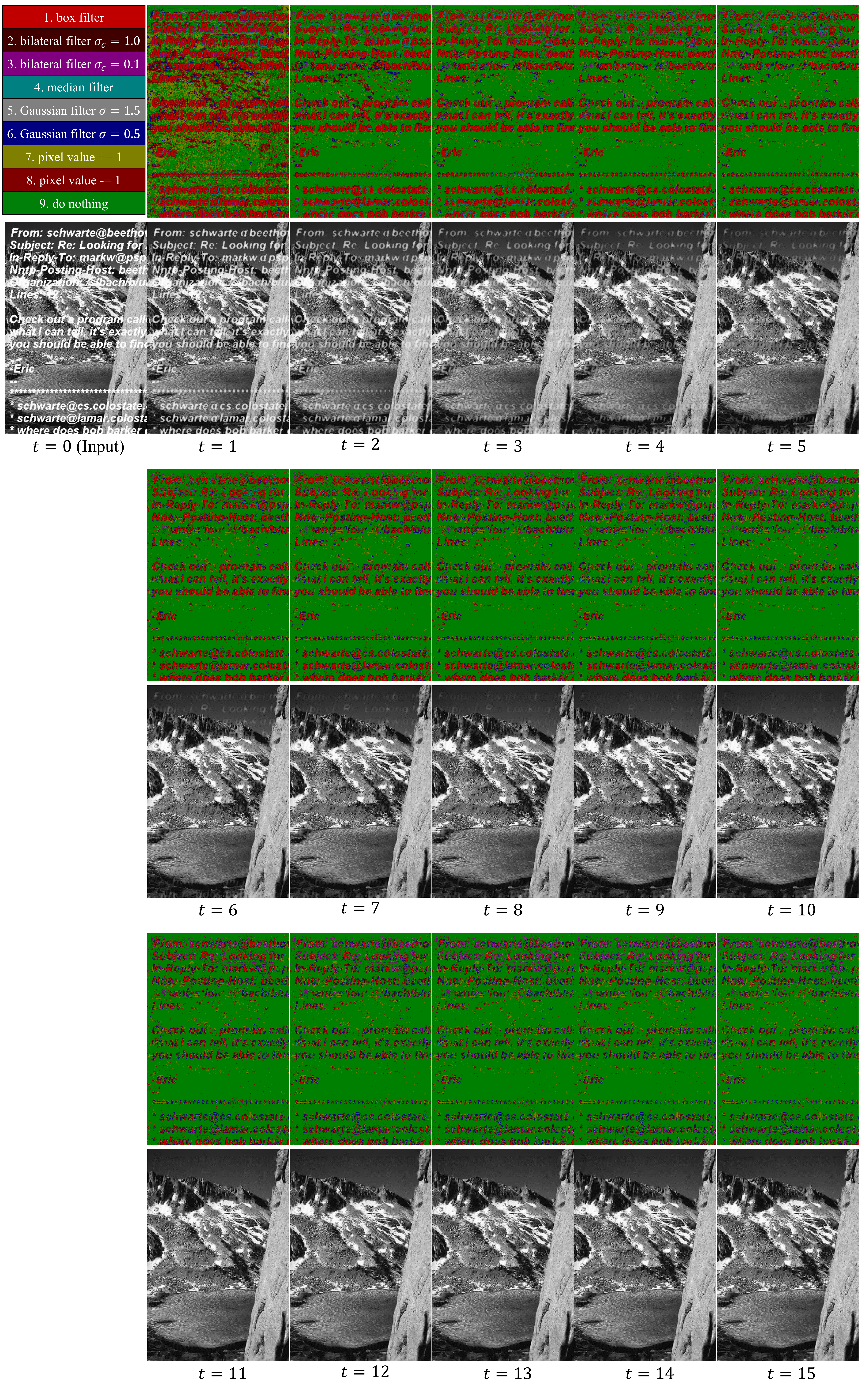}
        \caption{Restoration process of the proposed method and the action map at each time step.}
        \label{fig:vis_rest}
    \end{center}    
\end{figure}
\Fref{fig:vis_rest} is the visualization of restoration process of the proposed method, and the action map at each time step.
\begin{figure}[t]
    \begin{center}
        \includegraphics[width=1.0\linewidth]{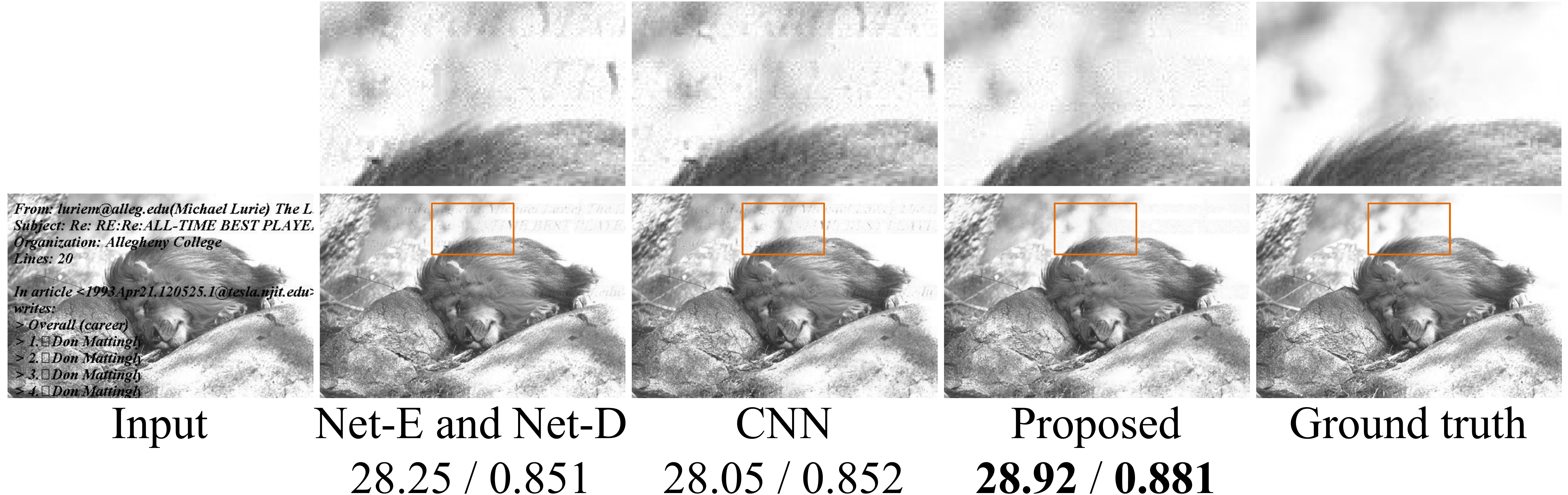}
        \caption{Qualitative comparison of the proposed method with Net-E and Net-D~\cite{DBLP:journals/corr/abs-1712-09078} and CNN~\cite{zhang2017learning} on image restoration. PSNR / SSIM are reported.}
        \label{fig:qualitative_rest}
    \end{center}    
\end{figure}
\Fref{fig:qualitative_rest} shows the qualitative comparison with Net-E and Net-D~\cite{DBLP:journals/corr/abs-1712-09078} and CNN~\cite{zhang2017learning}.
We observed that there are visually large differences between the results from the proposed method and those from the compared methods.

\subsection{Local Color Enhancement}\label{sec:color}
\subsubsection{Method}
\begin{table}[t]
\caption{Thirteen actions for local color enhancement.}
\centering
{
{\small
  \begin{tabular}{cccc} \toprule
      \multicolumn{3}{c}{Action} \\ \toprule
     1 / 2 & contrast & $\times0.95$ / $\times1.05$ \\
     3 / 4 & color saturation & $\times0.95$ / $\times1.05$ \\
     5 / 6 & brightness & $\times0.95$ / $\times1.05$ \\
     7 / 8 & red and green & $\times0.95$ / $\times1.05$ \\
     9 / 10 & green and blue & $\times0.95$ / $\times1.05$ \\
     11 / 12 & red and blue & $\times0.95$ / $\times1.05$ \\
     13 & do nothing & \\
 \bottomrule
  \end{tabular} 
}
}
\label{tbl:act_color}
\end{table}
We also applied the proposed method to the local color enhancement.
We used the dataset created by~\cite{yan2016automatic}, which has 70 train images and 45 test images downloaded from Flicker.
Using Photoshop, all the images were enhanced by a professional photographer for three different stylistic local effects: Foreground Pop-Out, Local Xpro, and Watercolor.
Inspired by~\cite{jongchan2018distort}, we decided the action set as shown in~\Tref{tbl:act_color}.
Given an input image $\bm{I}$, the proposed method changes the three channel pixel value at each pixel by executing an action.
When inputting $\bm{I}$ to the network, the RGB color values were converted to CIELab color values.
We defined the reward function as the decrease of L2 distance in the CIELab color space as follows:
\begin{equation}
r_i^{(t)}=|I_i^{target}-s_i^{(t)}|_2-|I_i^{target}-s_i^{(t+1)}|_2.\label{eq:reward_color}
\end{equation}
All the hyperparameters and settings were same as those in image restoration, except for the length of episodes, i.e., $t_{max}=10$.

\subsubsection{Results}
\begin{table}[t]
\caption{Comparison of mean L2 testing errors on local color enhancement. The errors except for the proposed method and pix2pix are from~\cite{yan2016automatic}.}
\centering
{
{\small
  \begin{tabular}{cc|ccc} \toprule
     \multicolumn{2}{c|}{\multirow{2}{*}{Method}} & Foreground & Local & \multirow{2}{*}{Watercolor} \\
      & & Pop-Out & Xpro &  \\ \toprule
     \multicolumn{2}{c|}{Original} & 13.86 & 19.71 & 15.30 \\
     \multicolumn{2}{c|}{Lasso} & 11.44 & 12.01 & 9.34 \\
     \multicolumn{2}{c|}{Random Forest} & 9.05 & 7.51 & 11.41 \\
     \multicolumn{2}{c|}{DNN~\cite{yan2016automatic}} & 7.08 & 7.43 & 7.20 \\
     \multicolumn{2}{c|}{Pix2pix~\cite{isola2017image}} & {\bf 5.85} & 6.56 & 8.84 \\
     \midrule
     \multicolumn{2}{c|}{Proposed} & & & \\
     +convGRU & +RMC & & & \\
      \checkmark& & 6.75 & 6.17 & 6.44 \\
      \checkmark&\checkmark & 6.69 & {\bf 5.67} & {\bf 6.41} \\
 \bottomrule
  \end{tabular} 
}
}
\label{tbl:comp_color}
\end{table}
\Tref{tbl:comp_color} shows the comparison of mean L2 errors on 45 test images.
The proposed method achieved better results than DNN~\cite{yan2016automatic} on all three enhancement styles, and comparable or slightly better results than pix2pix.
We observed that the RMC improved the performance although their degrees of improvement depended on the styles.
It is noteworthy that the existing color enhancement method using deep RL~\cite{jongchan2018distort,hu2017exposure} cannot be applied to this local enhancement application because they can execute only global actions.

\begin{figure*}[t]
    \begin{center}
        \includegraphics[width=0.85\linewidth]{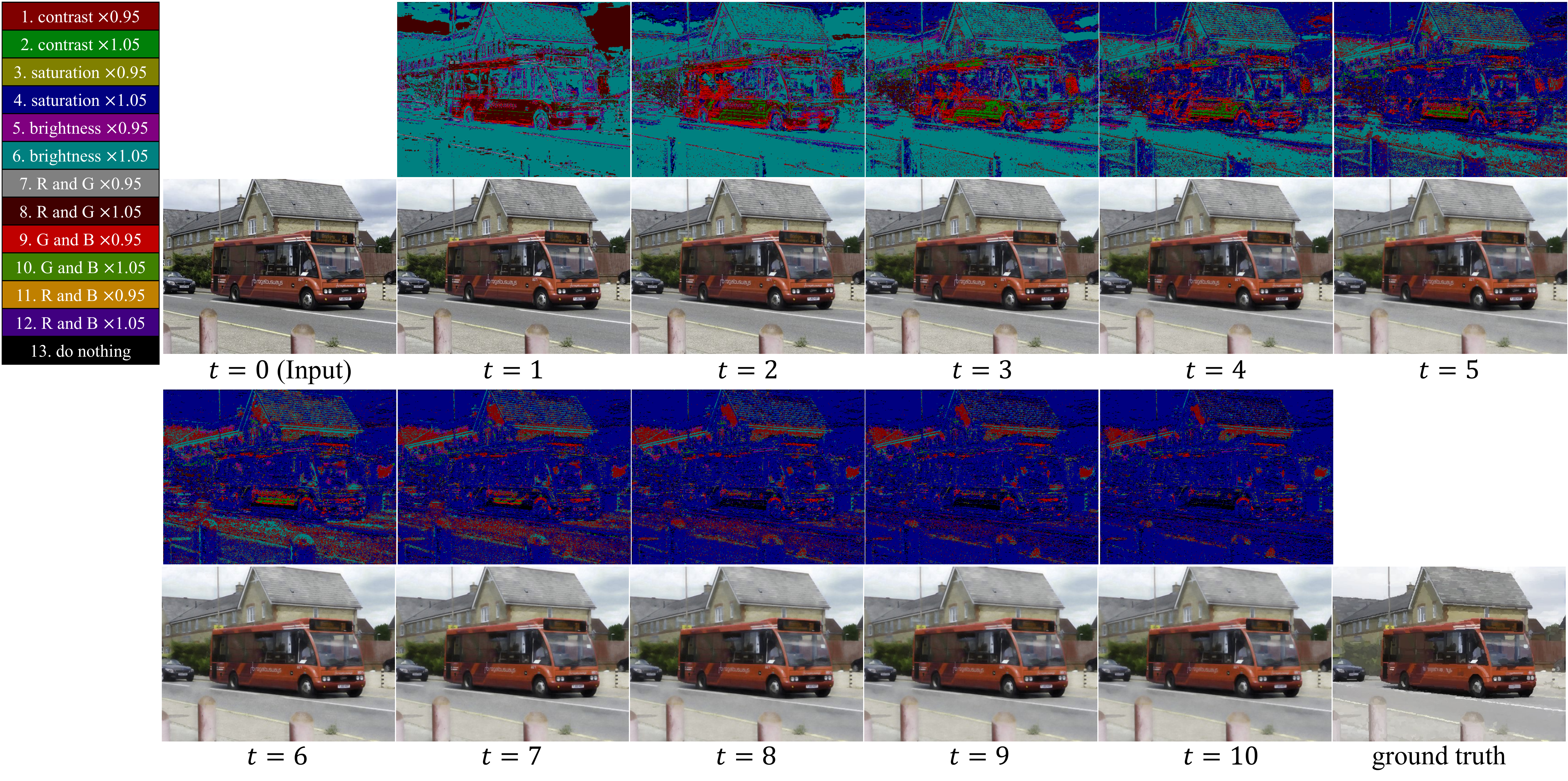}
        \caption{Color enhancement process of the proposed method for watercolor, and the action map at each time step.}
        \label{fig:vis_color}
    \end{center}    
\end{figure*}
\Fref{fig:vis_color} is the visualization of the color enhancement process of the proposed method, and the action map at each time step.
Similar to~\cite{jongchan2018distort}, the proposed method is interpretable while the DNN-based color mapping method~\cite{yan2016automatic} is not.
We can see that the brightness and saturation were mainly increased to convert the input image to watercolor style.

\begin{figure}[t]
    \begin{center}
        \includegraphics[width=1.0\linewidth]{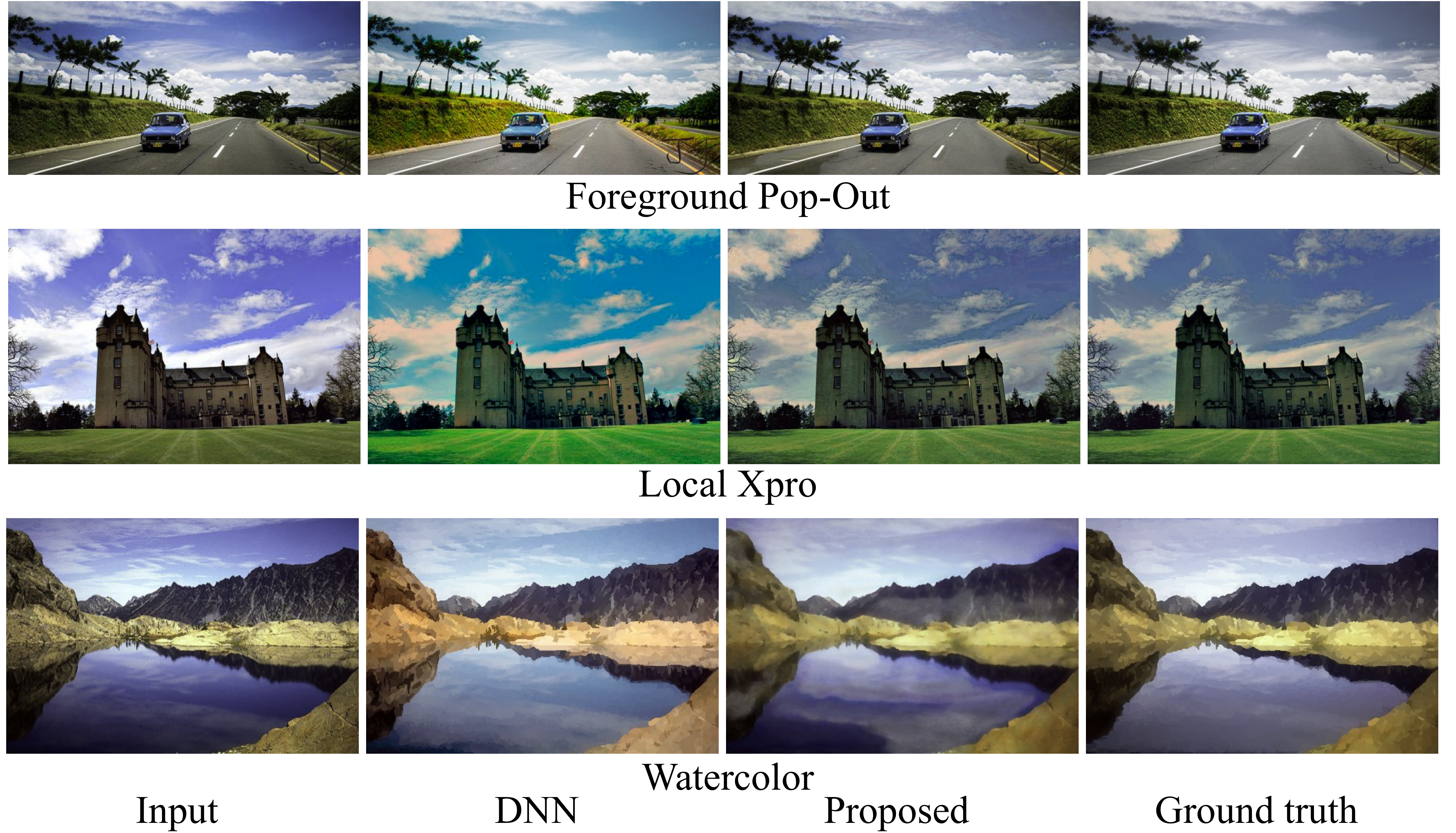}
        \caption{Qualitative comparison of the proposed method and DNN~\cite{yan2016automatic}. The saturation of the images from DNN appear higher owning to the color correction for the sRGB space (for details, see~\url{https://github.com/stephenyan1231/dl-image-enhance}).}
        \label{fig:qualitative_color}
    \end{center}    
\end{figure}
\Fref{fig:qualitative_color} shows the qualitative comparison between the proposed method and DNN~\cite{yan2016automatic}.
The proposed method achieved both quantitatively and qualitatively better results.

\section{Conclusions}
We proposed a novel pixelRL problem setting and applied it to three different applications: image denoising, image restoration, and local color enhancement.
We also proposed an effective learning method for the pixelRL problem, which boosts the performance of the pixelRL agents.
Our experimental results demonstrated that the proposed method achieved comparable or better results than state-of-the-art methods on each application.
Different from the existing deep learning-based methods for such applications, the proposed method is interpretable.
The interpretability of deep learning has been attracting much attention~\cite{selvaraju2017grad}, and it is especially important for some applications such as medical image processing~\cite{razzak2018deep}.

The proposed method can maximize the pixel-wise reward; in other words, it can minimize the pixel-wise non-differentiable objective function.
Therefore, we believe that the proposed method can be potentially used for more image processing applications where supervised learning cannot be applied.

\section{Acknowledgments}
This work was partially supported by the Grants-in-Aid for Scientific Research (no. 16J07267) from JSPS and JST-CREST (JPMJCR1686).

\newpage
\twocolumn[
\begin{center}
\textbf{\LARGE Supplementary Material}
\end{center}
]

\section{Network Architecture}
\begin{figure*}[t]
    \begin{center}
        \includegraphics[width=1.0\linewidth]{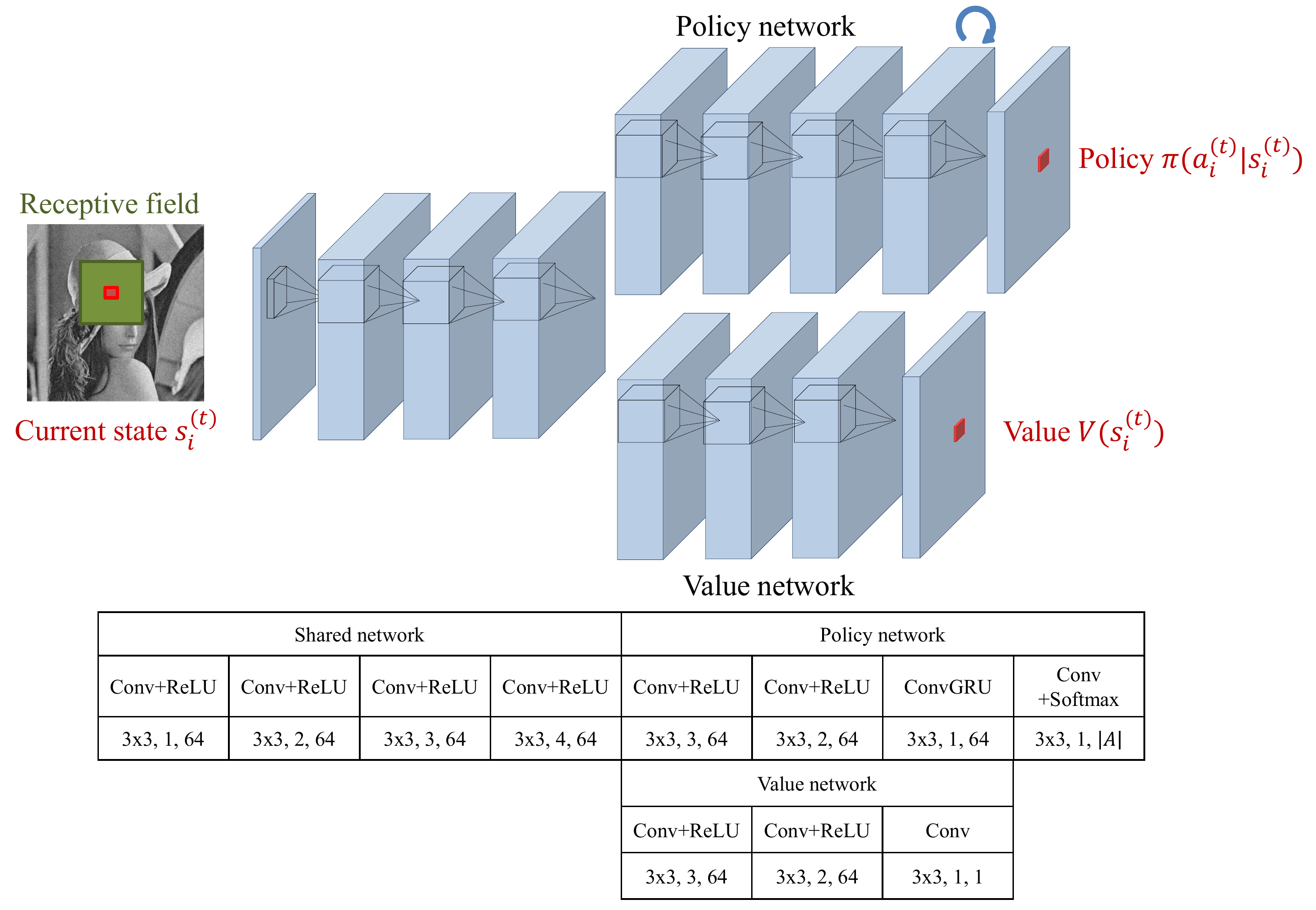}
        \caption{Network architecture of the fully convolutional A3C. The numbers in the table denote the filter size, dilation factor, and output channels, respectively.}
        \label{fig:arch}
    \end{center}    
\end{figure*}
Our architecture is illustrated in~\Fref{fig:arch}.

\section{Training Algorithm}
We summarize the training algorithm of the fully convolutional A3C with the proposed reward map convolution in Algorithm~\ref{alg}.
The differences from the original A3C are highlighted in red.
\begin{algorithm*}[!t]
\caption{Training pseudo-code of fully convolutional A3C with the proposed reward map convolution}
\label{alg}
\begin{algorithmic}
\STATE // Assume global shared parameter vectors $\theta_p$, $\theta_v$, and \textcolor{red}{$\bm{w}$} and global counter $T=0$.
\STATE // Assume thread-specific parameter vectors $\theta'_p$, $\theta'_v$, and \textcolor{red}{$\bm{w}'$}.
\STATE Initialize thread step counter $t\leftarrow1$.
\REPEAT
\STATE Reset gradients: $d\theta_p \leftarrow 0$, $d\theta_v \leftarrow 0$, and \textcolor{red}{$d\bm{w} \leftarrow 0$}.
\STATE Synchronize thread-specific parameters $\theta_p'=\theta_p$, $\theta_v'=\theta_v$, and \textcolor{red}{$\bm{w}'=\bm{w}$}
\STATE $t_{start}=t$
\STATE Obtain state $s_i^{(t)}$ for $\forall i$
\REPEAT
\STATE Perform $a_i^{(t)}$ according to policy $\pi(a_i^{(t)}|s_i^{(t)})$ for $\forall i$
\STATE Receive reward $r_i^{(t)}$ and new state $s_i^{(t+1)}$ for $\forall i$
\STATE $t\leftarrow t+1$
\STATE $T\leftarrow T+1$
\UNTIL terminal $s_i^{(t)}$ or $t-t_{start}==t_{max}$
\STATE for $\forall i$ $R_i=\begin{cases}0 & \text{for terminal}\ s_i^{(t)} \\ V(s_i^{(t)}) & \text{for non-terminal}\ s_i^{(t)}\end{cases}$
\FOR{$k \in \{t-1,\cdots,t_{start}\}$}
\STATE $R_i \leftarrow \gamma R_i$
\STATE \textcolor{red}{Convolve $\bm{R}$ with $\bm{w}$: $R_i \leftarrow \sum_{j\in \mathcal{N}(i)}w_{i-j}R_j$ for $\forall i$}
\STATE $R_i \leftarrow r_i^{(k)}+R_i$
\STATE Accumulate gradients w.r.t. $\theta'_p$: $d\theta_p\leftarrow d\theta_p-\nabla_{\theta_p'}\textcolor{red}{\frac{1}{N}\sum_{i=1}^N}\log\pi(a_i^{(k)}|s_i^{(k)})(R_i-V(s_i^{(k)}))$
\STATE Accumulate gradients w.r.t. $\theta'_v$: $d\theta_v\leftarrow d\theta_v+\nabla_{\theta_v'}\textcolor{red}{\frac{1}{N}\sum_{i=1}^N}(R_i-V(s_i^{(k)}))^2$
\STATE \textcolor{red}{Accumulate gradients w.r.t. $\bm{w}'$: $d\bm{w}\leftarrow d\bm{w}-\nabla_{\bm{w}}\frac{1}{N}\sum_{i=1}^N\log\pi(a_i^{(k)}|s_i^{(k)})(R_i-V(s_i^{(k)}))+\nabla_{\bm{w}}\frac{1}{N}\sum_{i=1}^N(R_i-V(s_i^{(k)}))^2$}
\ENDFOR
\STATE Update $\theta_p$, $\theta_v$, and \textcolor{red}{$\bm{w}$} using $d\theta_p$, $d\theta_v$, and \textcolor{red}{$d\bm{w}$}, respectively.
\UNTIL $T>T_{max}$
\end{algorithmic}
\end{algorithm*}

\section{Implementation Details}\label{sec:imple}
We set the minibatch size to 64, and the training images were augmented with $70\times 70$ random cropping, left-right flipping, and random rotation.
To train the fully convolutional A3C, we used ADAM optimizer~\cite{kingma2014adam} and the poly learning, where the learning rate started from $1.0\times 10^{-3}$ and multiplied by $(1-\frac{episode}{max\_episode})^{0.9})$ at each episode. 
We set the $max\_episode$ to 30,000 and the length of each episode $t_{max}$ to 5.
Therefore, the maximum global counter $T_{max}$ in Algorithm~\ref{alg} was $30,000 \times 5=150,000$.
To reduce the training time, we initialize the weights of the fully convolutional A3C with the publicly available weights of~\cite{zhang2017learning}, except for the convGRU and the last layers.
We adopted the stepwise training: the fully convolutional A3C was trained first, subsequently it was trained again with the reward map convolution.
We set the filter size of $\bm{w}$ to $33\times 33$, which is equal to the receptive field size of the networks in~\Fref{fig:arch}.
The number of asynchronous threads was one (i.e., equivalent to A2C: advantage actor-critic).
$\bm{w}$ was initialized as the identity mapping (i.e., only the center of $\bm{w}$ was one, and zero otherwise).
It required approximately 15.5 hours for the 30,000 episode training, and 0.44 sec on average for a test image whose size is $481\times 321$ on a single Tesla V100 GPU.

\section{Additional Experimental Results}
\subsection{Poisson Denoising}
\begin{table}[t]
\caption{PSNR [dB] on BSD68 test set with Poisson noise.}
\centering
{
  \begin{tabular}{ccc|ccc} \toprule
     \multicolumn{3}{c|}{\multirow{2}{*}{Method}} & \multicolumn{3}{c}{Peak Intensity}\\
      & & & 120 & 30 & 10 \\ \toprule
     \multicolumn{3}{c|}{CNN~\cite{zhang2017learning}} & 31.62 & 28.20 & 25.93 \\
     \multicolumn{3}{c|}{CNN~\cite{zhang2017learning} +aug.} & {\bf 31.66} & {\bf 28.26} & {\bf 25.96} \\ \midrule
     \multicolumn{3}{c|}{Proposed} & & & \\
     +convGRU & +RMC & +aug. & & & \\
      & & & 31.17 & 27.84 & 25.55 \\
      \checkmark& & & 31.28 & 27.94 & 25.64 \\
      \checkmark&\checkmark & & 31.37 & 27.95 & 25.70 \\
      \checkmark&\checkmark &\checkmark & 31.47 & 28.03 & 25.77 \\
 \bottomrule
  \end{tabular} 
}
\label{tbl:comp_poisson}
\end{table}
\Tref{tbl:comp_poisson} shows the comparison of Poisson denoising.
Similar to~\cite{luisier2011image}, we simulated the Poisson noise with different peak intensities.
The lower the peak intensity, the higher is the noise generated.
An almost similar tendency to Gaussian denoising was observed.
The proposed method achieved a slightly lower performance, compared with CNN~\cite{zhang2017learning}.

\section{Justification of Action Sets}
For local color enhancement, we simply chose the same action set as the prior work for global color enhancement~\cite{jongchan2018distort} except for the “do nothing” action. The motivation for the choice has been discussed in the section 4.1 in their paper.

\begin{table}[t]
\caption{PSNR [dB] on Gaussian denoising with different action sets.}
\centering
{
  \begin{tabular}{c|ccc} \toprule
     \multirow{2}{*}{Actions} & \multicolumn{3}{c}{std. $\sigma$}\\
      & 15 & 25 & 50 \\ \toprule
     only basic filters & 29.82 & 27.60 & 25.20 \\
     + bilateral filter $\sigma_c=1.0$ & 29.93 & 27.81 & 25.30 \\
     + pixel value $\pm=$ 1 & 30.72 & 28.25 & 25.59 \\
     + different filter parameters. & 31.26 & 28.83 & 25.87 \\
 \bottomrule
  \end{tabular} 
}
\label{tbl:ablation_gaussian}
\end{table}
For denoising, we conducted the ablation studies with different set of actions on Gaussian denoising. 
\Tref{tbl:ablation_gaussian} shows the results.
When the actions were only basic filters ([1] box filters, [4] median filter, [5] Gaussian filter with $\sigma=1.5$, and [9] do nothing in Table 1 in our main paper), its PSNRs were 29.82, 27.60, and 25.20 for noise std. 15, 25, and 50, respectively. 
When we added [2] bilateral filter with $\sigma_c=1.0$, the PSNRs increased to 29.93, 27.81, and 25.30. 
In addition, when we added the two actions ([7] pixel value += 1 and [8] pixel value -= 1), the PSNRs further increased to 30.72, 28.25, and 25.59. 
Finally, when we added the two filters with different parameters ([3] bilateral filter with $\sigma_c=0.1$ and [6] Gaussian filter with $\sigma=0.5$), the PSNRs were 31.26, 28.83, and 25.87. 
Therefore, all the actions are important for the high performance although it may be further increased if we can find more proper action set.
Although we tried adding some advanced filters for image denoising such as guided filter~\cite{he2010guided} and non-local means filter~\cite{buades2005non}, the performance was not improved any more.

\end{document}